\documentclass[11pt,a4paper]{article}
\usepackage[final]{arr}
\usepackage{times}
\usepackage{latexsym}
\usepackage{soul}
\usepackage{booktabs}
\usepackage{amsmath,caption}
\usepackage{colortbl}
\usepackage{arydshln}
\usepackage{multirow}
\usepackage{multicol}
\usepackage{hyperref}

\usepackage{microtype}
\usepackage{graphicx}
\usepackage{multirow}
\title{HiTab: A Hierarchical Table Dataset for Question Answering and Natural Language Generation}

\author{Zhoujun Cheng\textsuperscript{1}$\thanks{$^{*}$Equal contributions. Work done during Zhoujun and Zhiruo's internship at Microsoft Research Asia.}$,~Haoyu Dong\textsuperscript{2}$^{*}$$\thanks{$^{\dagger}$Corresponding author.}$~,~Zhiruo Wang\textsuperscript{3}$^{*}$,~Ran Jia\textsuperscript{2},~Jiaqi Guo\textsuperscript{4}       ,\\\textbf{Yan Gao}\textsuperscript{2},~\textbf{Shi Han}\textsuperscript{2},~\textbf{Jian-Guang Lou}\textsuperscript{2}, ~\textbf{Dongmei Zhang}\textsuperscript{2}
     \\
     \textsuperscript{1}Shanghai Jiao Tong University, ~\textsuperscript{2}Microsoft Research Asia \\
     \textsuperscript{3}Carnegie Mellon  University, ~\textsuperscript{4}Xi'an Jiaotong University\\
\tt{blankcheng@sjtu.edu.cn}, \tt{zhiruow@cs.cmu.edu} \\ \tt{jasperguo2013@stu.xjtu.edu.cn} \\
\tt {\{hadong,jia.ran,yan.gao,shihan,jlou,dongmeiz\}@microsoft.com}
}

\date{}

\begin{document}
\maketitle

\begin{abstract}

Tables are often created with hierarchies, but existing works on table reasoning mainly focus on flat tables and neglect hierarchical tables. Hierarchical tables challenge table reasoning by complex hierarchical indexing, as well as implicit relationships of calculation and semantics.
We present a new dataset, HiTab, to study question answering (QA) and natural language generation (NLG) over hierarchical tables.
HiTab is a cross-domain dataset constructed from a wealth of statistical reports and Wikipedia pages, and has unique characteristics: (1) nearly all tables are hierarchical, and (2) questions are not proposed by annotators from scratch, but are revised from real and meaningful sentences authored by analysts.
(3) To reveal complex numerical reasoning in analysis, we provide fine-grained annotations of quantity and entity alignment. 
%HiTab provides 10,686 descriptive sentences and QA pairs with well-annotated quantity and entity alignment on 3,597 tables with broad coverage of table hierarchies and numerical reasoning types.
Experimental results show that HiTab presents a strong challenge for existing baselines and a valuable benchmark for future research.
Targeting hierarchical structure, we devise an effective hierarchy-aware logical form for symbolic reasoning over tables. Furthermore, we leverage entity and quantity alignment to explore partially supervised training in QA and conditional generation in NLG, which largely reduces spurious predictions in QA and meaningless descriptions in NLG. The dataset and code are available at \url{https://github.com/microsoft/HiTab}.

%This dataset can new research into numerical reasoning over hierarchical tables.
%In the NLG task, we find that entity and quantity alignment helps NLG models to generate better results in a conditional generation setting. 

\end{abstract}

\section{Introduction} \label{sec:intro}

\begin{figure}[t]
    \begin{center}
    \includegraphics[width=3.in]{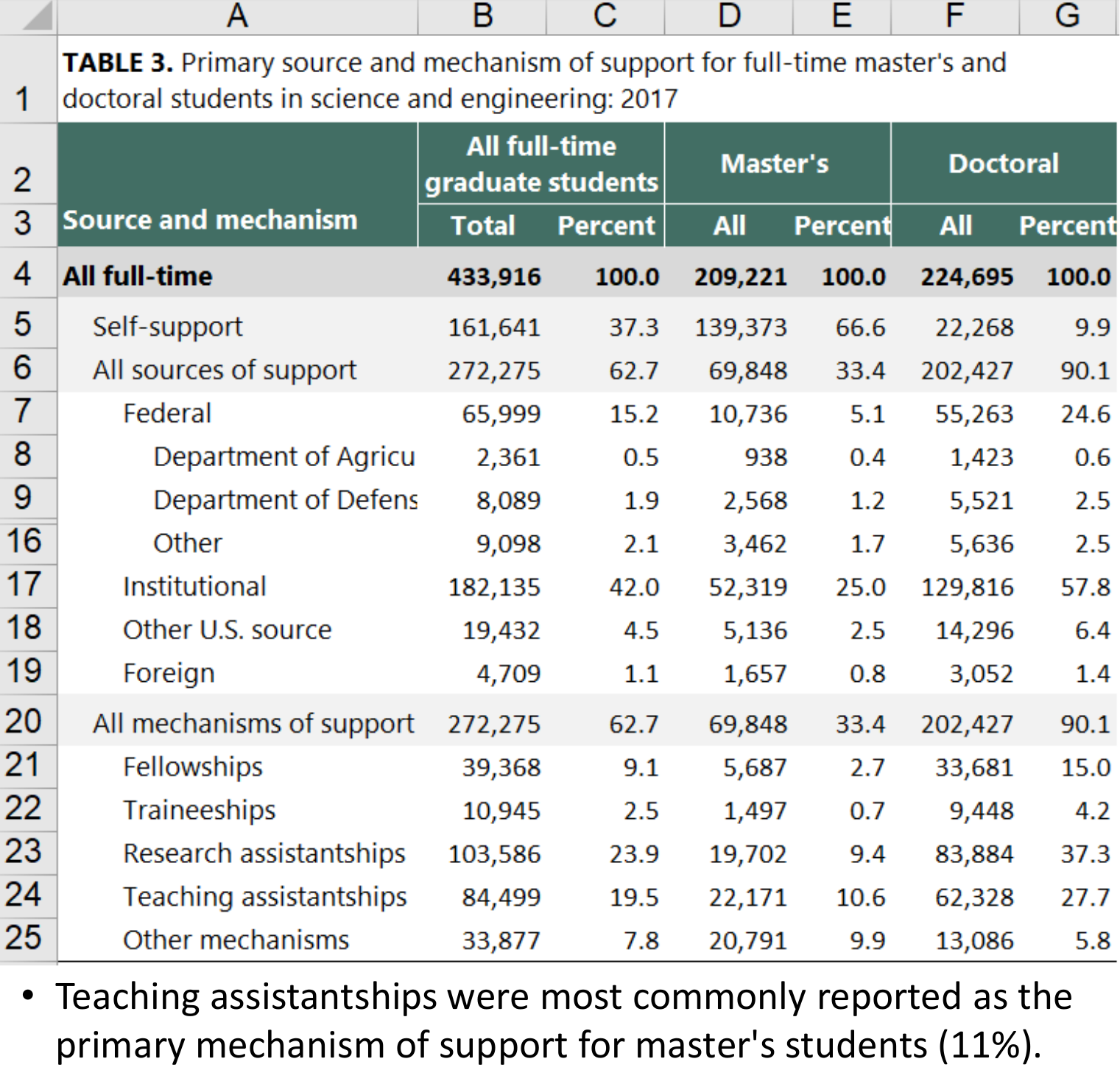}
    \end{center}
    \vspace{-0.3cm}
\caption{A hierarchical table and accompanied descriptions in a National Science Foundation report.\protect\footnotemark  }
\label{fig:realtable}
\vspace{-0.32cm}
\end{figure}
\footnotetext{https://www.nsf.gov/statistics/2019/nsf19319/}

In recent years, there are a flurry of works on reasoning over semi-structured tables, e.g., answering questions over tables~\cite{yu2018spider,pasupat2015compositional} and generating fluent and faithful text from tables~\cite{lebret2016neural,parikh2020totto}.
But they mainly focus on simple flat tables and neglect complex tables, e.g., hierarchical tables. A table is regarded as hierarchical if its header exhibits a multi-level structure~\cite{lim1999automated,chen2014integrating,wang2020structure}. 
Hierarchical tables are widely used, especially in data products, statistical reports, and research papers in government, finance, and science-related domains.

Hierarchical tables challenge QA and NLG due to: \textbf{(1) Hierarchical indexing.} Hierarchical headers, such as D2:G3 and A4:A25 in Figure~\ref{fig:realtable}, are informative and intuitive for readers, but make cell selection much more compositional than flat tables, requiring  multi-level and bi-dimensional indexing. For example, to select the cell E5 (``66.6''), one needs to specify two top header cells, ``Master's'' and ``Percent'', and two left header cells, ``All full-time'' and ``Self-support''. 
\textbf{(2) Implicit calculation relationships among quantities.} In hierarchical tables, it is common to insert aggregated rows and columns without explicit indications, e.g., total (columns B,D,F and rows 4,6,7,20) and proportion (columns C,E,G), which challenge precise numerical inference.
\textbf{(3) Implicit semantic relationships among entities.} There are various cross-row, cross-column, and cross-level entity relationships, but lack explicit indications, e.g., ``source'' and ``mechanism'' in A2 describe A6:A19 and A20:A25 respectively, and D2 (``Master's'') and F2 (``Doctoral'') can be jointly described by a virtual entity, ``Degree''. How to identify semantic relationships and link entities correctly is also a challenge.

In this paper, we aim to build a dataset for hierarchical table QA and NLG.
But without sufficient data analysts, it's hard to ensure questions and descriptions are meaningful and diverse~\cite{gururangan2018annotation,poliak2018hypothesis}. 
Fortunately, large amounts of statistical reports are public from a variety of organizations~\cite{statcan,nsf,census,cdc,bls,imf}, containing rich hierarchical tables and textual descriptions. 
Take Statistics Canada~\cite{statcan} for example, it consists of $6,039$ reports in $27$ domains authored by over 1,000 professionals.
Importantly, since both tables and sentences are authored by domain experts, sentences are natural and reflective of real understandings of tables.
%It inspires us to build target text for NLG and questions for QA based on existing descriptions instead of writing from scratch.
%It will not only save huge expert efforts, but also ensure target text and questions are meaningful, natural, and diverse.

To this end, we propose a new dataset, HiTab, for QA and NLG on hierarchical tables. \textbf{(1)} All sentence descriptions of hierarchical tables are carefully extracted and revised by human annotators. \textbf{(2)} 
It shows that annotations of fine-grained and lexical-level entity linking significantly help table QA~\cite{lei2020re,shi2020potential}, motivating us to align entities in text with table cells. In addition to entity, we believe aligning quantities~\cite{ibrahim2019bridging}, especially composite quantities (computed by multiple cells), is also important for table reasoning, so we annotate underlying numerical relationships between quantities in text and table cells, as Table~\ref{tab:annotation} shows. 
%We believe that entity alignment~\cite{jimenez2020semtab} and quantity alignment~\cite{ibrahim2019bridging} are not only two important tasks in themselves, but also generic and helpful for various tasks requiring table-text joint understanding.
\textbf{(3)} Since real sentences in statistical reports are natural, diverse, and reflective of real understandings of tables, we devise a process to construct QA pairs based on existing sentence descriptions instead of asking annotators to propose questions from scratch.
%. Annotators convert sentence descriptions to question-answering pairs and use spreadsheet formulas to record the calculation process of answering

HiTab presents a strong challenge to state-of-the-art baselines. For the QA task, MAPO~\cite{mapo} only achieves $29.2\%$ accuracy due to the ineffectiveness of the logical form customized for flat tables.
To leverage the hierarchy for table reasoning, we devise a hierarchy-aware logical form for table QA, which shows high effectiveness. We propose partially supervised training given annotations of linked mentions and formulas, which helps models to largely reduce spurious predictions and achieve $45.1\%$ accuracy.
For the NLG task, models also have difficulties in understanding deep hierarchies and generate complex analytical texts. We explore controlled generation~\cite{parikh2020totto}, showing that conditioning on both aligned cells and calculation types helps models to generate meaningful texts.

\begin{table*}[t]
\vspace{-0.5cm}
    \begin{center}
    \includegraphics[width=6.1in]{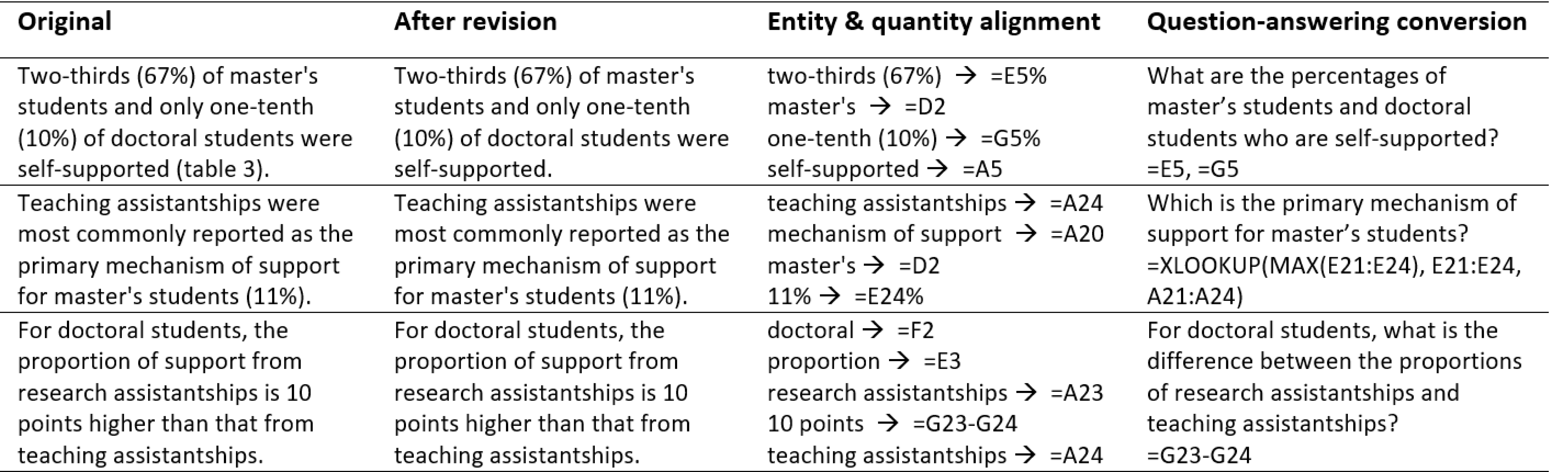}
    \end{center}
\vspace{-0.3cm}
\caption{Examples of the annotation process. All sentences describe the table in Figure~\ref{fig:realtable}.}
\label{tab:annotation}
\vspace{-0.32cm}
\end{table*}

\section{Dataset Construction and Analysis }\label{sec:dataset}
% 
 %Each student is paid \$ 7.8 an hour, and they totally spend 2,400 hours.
 %We recruit 18 students or graduates (13 females and 5 males) in computer science, finance, and English majors from top universities. 
We design an annotation process with six steps. To well-handle the annotation complexity, we recruit 18 students or graduates (13 females and 5 males) in computer science, finance, and English majors from top universities, and provide them with comprehensive online training, documents, and QAs. The annotation totally costs 2,400 working hours. We will discuss the ethical considerations in Section~\ref{appendix:ethics}.

\subsection{Hierarchical Table Collection}
%A large number of reports from various organizations are publicly available.
We select two representative organizations, Statistics Canada~\cite{statcan} and National Science Foundation~\cite{nsf}, that are rich of statistical reports. Different from ~\citet{census,cdc,bls,imf} that only provide PDF reports where table hierarchies are hard to extract precisely~\cite{schreiber2017deepdesrt}, StaCan and NSF also provide reports in HTML, from which cell information such as text and formats can be extracted precisely using HTML tags. 

First, we crawl English HTML statistical reports published in recent five years from StatCan ($1,083$ reports in $27$ well-categorized domains) and NSF ($208$ reports from $11$ organizations in science foundation domain). We merge StatCan and NSF and get the combination of various domains. In addition, ToTTo contains a small proportion ($5.03\%$) of hierarchical tables, so we include them to cover more domains from Wikipedia. To keep the balance between statistical reports and Wikipedia pages, we include random $1,851$ tables~($50\%$ of our dataset) from ToTTo. Next, we transform HTML tables to spreadsheet tables using a preprocessing script. Since spreadsheet formula is easy to write, execute, and check, the spreadsheet is naturally a great annotation tool to align quantities and answer questions. To enable correct formula execution, we normalize quantities in data cells by excluding surrounding superscripts, internal commas, etc. Extremely small or large tables are filtered out (Appendix~\ref{appendix:dataprepare} gives more details).

\subsection{Sentence Extraction and Revision}
In this step, annotators manually go through statistical reports and extract sentence descriptions for each table. 
Sentences consisting of multiple semantic-independent sub-sentences will be carefully split into multiple ones.
Annotators are instructed to eliminate redundancy and ambiguity in sentences through revisions including decontextualization and phrase deletion~\cite{parikh2020totto}. Fortunately, most sentences in statistical reports are clean and fully supported by table data, so few revisions are needed to get high-quality text.

\begin{table}[h]
\vspace{-0.15cm}
    \scalebox{0.65}{
    \begin{tabular}{l l}
    	\hline
    	\textbf{Operators}& \textbf{Formula template (ranges are placeholders)}  \\
    	\hline
    	{opposite}, {percent} & {=-A5}, {=B2\%} \\
    	\hline
    	{kth-argmax/argmin} & {=XLOOKUP(SMALL(D1:D3, k), D1:D3, A1:A3)}. \\
    	{pair-argmax/argmin} & {=IF(B1$>$B2, A1, A2)}\protect\footnotemark  \\
    	\hline
    	{sum}, {average} & {=SUM(D2:D4)}, {=AVERAGE(D2:D4)} \\
    	{max}, {count} & {=MAX(D2:D4)}, {=COUNT(D2:D4)} \\
    	{diff}, {div} & {=D3-D4}, {=D3/D4}\\
    	 
    	\hline
    \end{tabular}
    }
    \vspace{-0.3cm}
    \caption{Example operators and formula templates. }
    \vspace{-0.3cm}
    \label{tab:formulas}
    \end{table}
\footnotetext{For samples with XLOOKUP or IF formulas, we didn't explicitly provide the formulas in dataset because some reasoning logics are still too complex to be covered by them, e.g., the candidate cells are not on a continuous row/column. Instead, we manually check the answer cell(s) and provide the answer cell reference(s) for these samples.}

\subsection{Entity and Quantity Alignment}
In this phase, annotators are instructed to align mentions in text with corresponding cells in tables. It has two parts, entity alignment and quantity alignment, as shown in Table~\ref{tab:annotation}. For entity alignment, we record the mappings from entity mentions in text to corresponding cells. 
Single-cell quantity mentions can be linked similar with entity mentions, but composite quantity mentions are calculated from two or more cells through operators like \textit{max/sum/div/diff} (Table~\ref{tab:formulas}).
The spreadsheet formula is powerful and easy-to-use for tabular data calculation, so we use the formula to record the calculations process of composite quantities in text, e.g., `10 points higher' (\textit{=G23-G24}). Although quantities are often rounded in descriptions, we neglect rounding and refer to precise quantities in table cells. 

\begin{table*}
\vspace{-0.5cm}
\scalebox{0.54}{
    \begin{tabular}{l r | l l c| c c | c c c c c}
    \hline
    \multirow{3}{*}{\textbf{Dataset}} &
    \multirow{3}{*}{\textbf{\# Tables}} &
    \multicolumn{3}{c|}{\textbf{Data source}} & \multicolumn{2}{c|}{\textbf{Fine-grained alignment}} &
    \multicolumn{5}{c}{\textbf{QA and NLG tasks}}
    % \multicolumn{6}{c}{\textbf{QA and NLG tasks}}
    
    \\ &\multirow{1}{*}{}
    &\multirow{2}{*}{Table} 
    &\multirow{1}{*}{Question}
    &\multirow{1}{*}{Real sentences}
    
     &\multirow{2}{*}{Entity}
     &\multirow{2}{*}{Quantity} 
     &\multirow{2}{*}{QA}
    &\multirow{2}{*}{NLG}
    &\multirow{2}{*}{Questions}
    &\multirow{1}{*}{Words per}
    &\multirow{2}{*}{Sentences} 
    %  &\multirow{1}{*}{Words per}
    \\
    &\multicolumn{1}{c}{}
    &\multicolumn{1}{|c}{}
    &\multirow{1}{*}{or sentence}
    &\multirow{1}{*}{revised per table}
     &\multirow{1}{*}{} 
     &\multirow{1}{*}{} 
     &\multirow{1}{*}{} 
     &\multirow{1}{*}{} 
     &\multirow{1}{*}{} 
     &\multirow{1}{*}{question} 
     &\multirow{1}{*}{} 
    % &\multirow{1}{*}{sentence} 
    \\
    
    \hline
    
    WTQ~\cite{pasupat2015compositional} & 2,108 & Wikipedia & Post-created & - & - & - & Yes & - & 22,033 & 10.0 & -  \\ % &  -    \\
    WikiSQL~\cite{zhong2017seq2sql}  & 26,521 & Wikipedia & Post-created & - & - & - & Yes & -& 80,654 & 11.7 & - \\ % &  -  \\
    Spider~\cite{yu2018spider} &  1,020  & College data,WikiSQL & Post-created & - & - & - & Yes & -  & 10,181 & 13.2 & - \\ %  & - \\
    %\hline
    HybridQA~\cite{chen2020hybridqa} & 13,000 & Wikipedia & Post-created & - & - & - & Yes & -  & 69,611 & 18.9 & - \\ %  &  - \\
    TAT-QA~\cite{zhu2021tat} & 2,757 & Financial reports (PDF) & Post-created & - & - & - & Yes & -  & 16,552 & 12.5 & -  \\ %  & -  \\
    FinQA~\cite{chen2021finqa} & 2,776 & Financial reports (PDF) & Post-created & - & - & - & Yes & -  & 8,281 & 16.6 & -  \\ %  & -  \\
    
    DART~\cite{nan2020dart} & 5,623 & WTQ,WikiSQL,... & Post-created & - & - & - & - & Yes  & - & - & 82,191 \\ %  & 19.6\\
    %SciGen & 547 & Scientific papers (CS) & Pre-existing &  - & - & - & Yes & - & 1,338 & - & -  \\
    %WebNLG,E2E
    LogicNLG~\cite{chen2020logical} & 7,392 & Wikipedia & Post-created & - & - & - & - & Yes  & - & - & 37,015 \\ %  & 13.9  \\
    %\hline
    ToTTo~\cite{parikh2020totto} &  83,141 & Wikipedia & Pre-existing & 1.4 & - & - & - & Yes & - & - & 120,000 \\ %  & 14.9  \\
    NumericNLG~\cite{suadaa2021towards} & 1,300 & Scientific papers (ACL) & Pre-existing &  3.8 & - &  - & - & Yes  & - & - & 4,756 \\ % 20.3  \\
    \textbf{HiTab} & \textbf{3,597} & \textbf{Stat. reports, Wiki.} & \textbf{Pre-existing} & \textbf{5.0 (reports)}  & \textbf{Yes} & \textbf{Yes} & \textbf{Yes} & \textbf{Yes} & \textbf{10,672} & \textbf{16.5} & \textbf{10,672} \\ % & \textbf{16.0}  \\
    \hline

    \end{tabular}

}
\vspace{-0.2cm}
\caption{Dataset statistics and comparison.}\label{tab:stat}
\end{table*}

\subsection{Converting Sentences to QA Pairs}
Existing QA datasets instruct annotators to propose questions from scratch, but it's hard to guarantee the meaningfulness and diversity of proposed questions. In HiTab, we simply revise declarative sentences into QA pairs. For each sentence, annotators need to identify a target key part to question about (according to the underlying logic), then convert it to the QA form. All questions are answered by formulas that reflect the numerical inference process. For example, the `XLOOKUP' operator is frequently used to retrieve the header cells of superlatives, as shown in Table~\ref{tab:annotation}. To keep sentences as natural as they are, we do not encourage unnecessary sentence modification during the conversion. If an annotator finds multiple ways to question regarding a sentence, he/she only needs to choose one way that best reflects the overall meaning.

\subsection{Regular Inspections and the Final Review}
We ask the two most experienced annotators to perform regular inspections and the final review. (1) In the labeling process, they regularly sample annotations (about $10\%$)  from all annotators to give timely feedback on labeling issues. (2) Finally, they review all annotations and fix labeling errors. Also, to assist the final review, we write a script to automatically identify spelling issues and formula issues.
To double-check the labeling quality before the final review, we study the agreement of annotators by collecting and comparing annotations on randomly sampled $50$ tables from two annotators.
It shows $0.89$ and $0.82$ for quantity and entity alignment in Fleiss Kappa respectively, which are regarded as ``almost perfect agreement''~\cite{landis1977measurement}, and $64.5$ in BLEU-4 after sentence revision, which also indicates high agreement. We further show annotation artifacts are substantially avoided in our dataset in Appendix~\ref{appendix:annotation artifacts}.

\subsection{Hierarchy Extraction} \label{subsec:hierarchy extraction}
We follow existing work ~\cite{lim1999automated,chen2014integrating,wang2020structure} and use the tree structure to model hierarchical headers.
Since cell formats such as merging, indentation, and font bold, are commonly used to present hierarchies, we adapt heuristics in~\cite{wang2020structure} to extract top and left hierarchical trees, which has high accuracy. We go through $100$ randomly sampled tables in HiTab,  $94\%$ of them are precisely extracted. Figure ~\ref{fig:linearization} in Appendix shows an illustration.

\begin{figure}[t]
\vspace{-0.3cm}
    \begin{center}
    \includegraphics[width=3.in]{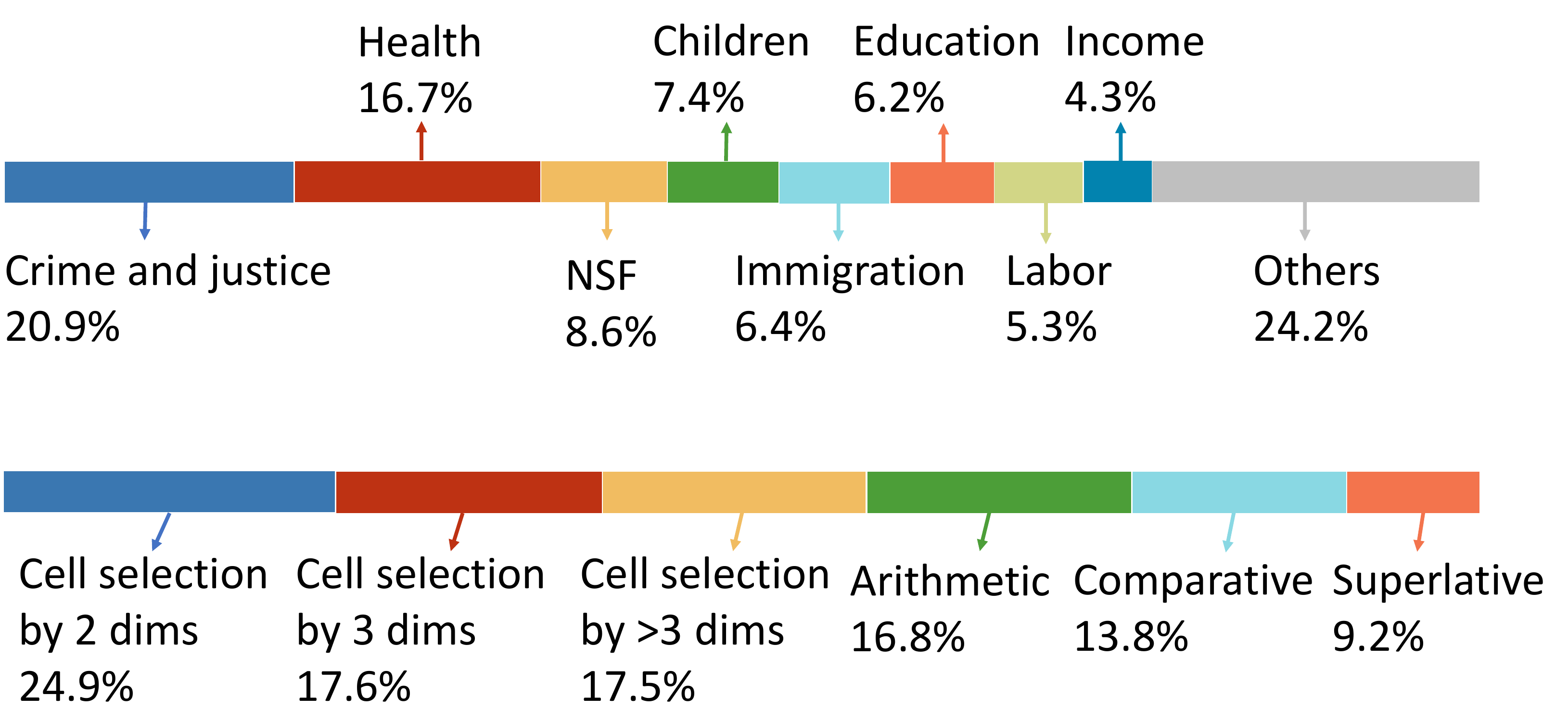}
    \end{center}
\vspace{-0.3cm}
\caption{Distribution of domains and operations in StatCan and NSF. \textit{Cell selection by k dims} means that header cells in \textit{k} levels are used in cell selection.}
\label{fig:distribution}
\vspace{-0.18cm}
\end{figure}

\subsection{Dataset Statistics and Comparison}
Table~\ref{tab:stat} shows a comprehensive comparison of related datasets.
%including WTQ~\cite{pasupat2015compositional}, WikiSQL~\cite{zhong2017seq2sql}, Spider~\cite{yu2018spider}, TAT-QA~\cite{zhu2021tat}, HybridQA~\cite{chen2020hybridqa}, FinQA~\cite{chen2021finqa}, DART~\cite{nan2020dart}, LogicalNLG~\cite{chen2020logical}, NumeicNLG~\cite{suadaa2021towards}, and ToTTo~\cite{parikh2020totto}. 
HiTab is not among the largest ones, but \textbf{(1)} it is the first dataset to study QA and NLG over hierarchical tables (accounting for 98.1\% tables in HiTab) in-depth;
\textbf{(2)} it is annotated with fine-grained entity and quantity alignment;
\textbf{(3)} compared with TAT-QA, FinQA, and NumericNLG that are single-domain, HiTab has a wide coverage of different domains from statistical reports and Wikipedia, even wider than ToTTo or WTQ that only involves Wikipedia tables;
\textbf{(4)} the number of real descriptions per table ($5.0$) in statistical reports (HiTab) is much richer than $1.4$ in Wikipedia (ToTTo) and $3.8$ in scientific papers, contributing more analytical aspects per table.
%; the average length of questions ($16.5$) in HiTab is longer than question lengths in most datasets where questions are post-created. 

Figure \ref{fig:distribution} analyzes this dataset by domains and operations: domains are diverse, covering $28$ domains from statistical reports (fully listed in Appendix~\ref{appendix:domain distribution}) and other open domains from Wikipedia; a large proportion of questions involves complex cell selection and numerical operations.

\section{Hierarchical Table QA}
Table QA is essential for table understanding, document retrieval, ad-hoc search, \textit{etc}. Hierarchical tables are quite common in these scenarios like in webpages and reports, while current Table QA tasks and methods focus on simple flat tables.
\vspace{-0.1cm}
\paragraph{Problem Statement} Hierarchical Table QA is defined as follows: given a hierarchical table $t$ and a question $x$  in natural language, output answer $y$. The question-answer pair should be fully supported by the table. Our dataset $D=\{(x_i,t_i,y_i)\}, i \in [1, N]$ is a set of $N$ question-table-answer triples. 

Table QA is usually formulated as a semantic parsing problem~\cite{pasupat2015compositional,liang2017neural}, where a parser converts the question into logical form, and an executor executes it to produce the answer. However, existing logical forms for Table QA~\cite{pasupat2015compositional,liang2017neural,yin2020tabert} are customized for flat or database tables. The three challenges mentioned in Section \ref{sec:intro} (hierarchical indexing, implicit indexing relationships, and implicit semantic relationships) make QA more difficult on hierarchical tables.

% ---------------------------------------

\subsection{Hierarchy-aware Logical Forms}
To this end, we propose a hierarchy-aware logical form that exploits table hierarchies to mitigate these challenges. Specifically, we define \textit{region} as the operating object, and propose two functions for hierarchical region selection.
\vspace{-0.1cm}
\paragraph{Definitions} Given tree hierarchies of tables extracted in Section~\ref{subsec:hierarchy extraction}, we define \textit{header} as a header cell (e.g., A7(``Federal'') in Figure \ref{fig:realtable}), and \textit{level} as a level in the left/top tree (e.g., A5,A6,A20 are on the same level). Existing logical forms on tables treat rows as operating objects and columns as attributes, and thus can not perform arithmetic operations on cells in the same row. However, a row in hierarchical tables is not necessarily a subject or record, thus operations can be applied on cells in the same row. Motivated by this, we define \textit{region} as our operating object, which is a data region in table indexed by both left and top headers (e.g., B6:C19 is a rectangular region indexed by A6,B2). The logical form execution process is divided into two phases: region selection and region operation.
\vspace{-0.1cm}
\paragraph{Region Selection} We design two functions $(filter\_tree\;h)$ and  $(filter\_level\;l)$ to do region selection, where $h$ is a header, $l$ is a level. Functions can be applied sequentially: the subsequent function applies on the return region of the previous function. $(filter\_tree\;h)$ selects a sub-tree region according to a header cell $h$: if $h$ is a leaf header (e.g., A8), the selected region should be the row/column indexed by $h$ (row 8); if $h$ is a non-leaf header (e.g., A7), the selected region should be the rows/columns indexed by both $h$ and its children headers (row 7-16). $(filter\_level\;l)$ selects a sub-tree from the input tree according to a level $l$ and return the sub-region indexed by headers on level $l$. These two functions mitigate aforementioned three challenges: (1) hierarchical indexing is achieved by applying these two functions sequentially; (2) with $filter\_level$, data with different calculation types (e.g., rows 4-5) will not be co-selected, thus not incorrectly operated together; (3) level-wise semantics can be captured by aggregating header cell semantics (e.g., embeddings) on this level. Some logical form execution examples are shown in Appendix~\ref{appendix:qa examples}.
\vspace{-0.1cm}
\paragraph{Region Operation}
Operators are applied on the selected region to produce the answer. We define $19$ operators, mostly following MAPO~\cite{mapo}, and further include some operators (e.g., \textit{difference rate}) for hierarchical tables. Complete logical form functions are shown in Appendix~\ref{appendix:logical form}.
% Operators are applied on the selected region to produce the answer. Simple cell selection (no operator), single operator, and composite operators are all allowed. We define $19$ operators, mostly following MAPO~\cite{mapo}, and further include some operators (e.g. \textit{difference rate}) for hierarchical tables. Complete logical form functions are shown in Appendix~\ref{appendix:logical form}. 

\subsection{Experimental Setup}
\subsubsection{Baselines}
We present baselines in two branches. One is logical form-based semantic parsing, and the other is end-to-end table parsing without logical forms.

\noindent \textbf{Neural Symbolic Machine}~\cite{liang2017neural} is a powerful semantic parsing framework consisting of a programmer to generate programs from NL and save intermediate results, and a computer to execute programs. We replace the LSTM encoder with BERT~\cite{devlin2018bert}, and implement a lisp interpreter for our logical forms as executor. Table is linearized by placing headers in level order, which is shown in detail in Appendix~\ref{appendix:linearization}. 

% Note that we do not consider TaBERT~\cite{yin2020tabert} as baseline encoder because (1) its core mechanisms are customized for flat tables, (2) its linearization is optimized for capturing column embedding, which is coupled with downstream logical forms (\textit{i.e.} columns are candidate output of decoding steps). Thus, we consider vanilla BERT as the baseline encoder.

\noindent \textbf{TaPas}~\cite{herzig2020tapas} is a state-of-the-art end-to-end table parsing model without generating logical forms. Its power to select cells and reason over tables is gained from its pretraining on millions of tables. To fit TaPas input, we convert hierarchical tables into flat ones following WTQ~\cite{pasupat2015compositional}. Specifically, we unmerge the cells spanning many rows/columns on left/top headers and duplicate the contents into unmerged cells. The first top header row is specified as column names. 
%Specifically, by unmerging merged cells in the top header, ``flattening'' top headers by column-wise concatenation, and specifying flattened top headers as column names. 

\subsubsection{Weak Supervision}
In weak supervision, the model is trained with QA pairs, without golden logical forms. For NSM, we compare three widely-studied learning paradigms:

    % \noindent \textbf{MML}~\cite{dempster1977maximum} maximizes marginal likelihood of observed programs.
    
    % \noindent \textbf{REINFORCE}~\cite{williams1992simple} maximizes the reward of on-policy samples.
    
    % \noindent \textbf{MAPO}~\cite{mapo} learns from programs both inside and outside the buffer, and samples with high efficiency by systematic exploration.
    
    \textbf{MML}~\cite{dempster1977maximum} maximizes the marginal likelihood of observed programs. \textbf{REINFORCE}~\cite{williams1992simple} maximizes the reward of on-policy samples. \textbf{MAPO}~\cite{mapo} learns from programs both inside and outside buffer, and samples efficiently by systematic exploration.
    
Since these methods require consistent programs for learning or warm start, we randomly search $15,000$ programs per sample before training. The pruning rules are shown in Appendix~\ref{appendix:trigger words}. Finally, $6.12$ consistent programs are found per sample.
    
% MML needs to learn from consistent programs, \textit{i.e.} programs that produce correct answers. REINFORCE and MAPO need consistent programs for warm up. Thus we randomly search $300$ iterations (\textasciitilde$15000$ programs per sample) for training set. We apply pruning rules shown in Appendix~\ref{} following ~\citet{mapo} in searching.  Finally, $6.12$ consistent programs are found for each sample.

For TaPas, we use the pre-trained version and follow its weak supervised training process on WTQ.

\subsubsection{Partial Supervision}
Given labeled entity links, quantity links, and calculations (from the formula), we further explore to guide training in a \textit{partially supervised} way. These three annotations indicate selected headers, region, and operators in QA\footnote{Entity and quantity alignments in text also occur in the question in most cases. In QA, we apply a simple n-gram matching algorithm to filter out the alignments not in questions.}. For NSM, we exploit them to prune spurious programs, \textit{i.e.}, incorrect programs that accidentally produce correct answers, in two ways. (1) When searching consistent programs, besides producing correct answers, programs are required to satisfy at least two constraints. In this way, the average consistent programs reduces from $6.12$ to $2.13$ per sample. (2) When training, satisfying each condition will add $0.2$ to the original binary 0/1 reward. Sampled programs with reward $r\geq1.4$ are added to the program buffer.

For TaPas, we additionally provide answer coordinates and calculation types in training following its WikiSQL setting.

\subsubsection{Evaluation Metrics}
We use \textit{Execution Accuracy} (\textit{EA}) as our metric following \cite{pasupat2015compositional}, measuring the percentage of samples with correct answers. We also report \textit{Spurious Program Rate} to study the percentage that incorrect logical forms produce correct answer. Since we do not have golden logical forms, we manually annotate logical forms for $150$ random samples in dev set for evaluation. 

\subsubsection{Implementations}
We split $3,597$ tables into train ($70\%$), dev ($15\%$) and test ($15\%$) with no overlap. We download pre-trained models from huggingface~\footnote{https://huggingface.co/}. For NSM, we utilize `bert-base-uncased', and fine-tune $20$K steps on HiTab. Beam size is $5$ for both training and inference. To test MAPO original logical form, we convert flatten tables as we do for TaPas. For TaPas, we adopt the PyTorch~\cite{paszke2019pytorch} version in huggingface. We utilize `tapas-base', and fine-tune $40$ epochs on HiTab. All experiments are conducted on a server with four V100 GPUs.

% ---------------------------------------
\begin{table}[t]
\small
\scalebox{0.86}{
\begin{tabular}{l c c c}
\toprule[1.2pt]
\multicolumn{4}{c}{\textit{Weak Supervision}} \\
\textbf{Method}  & \textbf{Dev} & \textbf{Test} & \textbf{\%Spurious}\\
\hline
MAPO $w.$ original logical form & $31.9$ & $29.2$ & - \\
TaPas $w/o.$ logical form & $39.7$ & $38.9$ & - \\
MML $w.$ h.a. logical form & $38.9$ & $36.7$ & $22.7$  \\
REINFORCE $w.$ h.a. logical form & $42.7$ & $38.4$ & $39.3$ \\
MAPO $w.$ h.a. logical form & $\textbf{43.5}$ & $\textbf{40.7}$ & $\textbf{19.0}$ \\
\hline
\multicolumn{4}{c}{\textit{Partial Supervision}} \\
TaPas $w/o.$ logical form & $41.2$ & $40.1$ & - \\
MML $w.$ h.a. logical form & $\textbf{45.4}$ & $\textbf{45.1}$ & $\textbf{10.3}$ \\
REINFORCE $w.$ h.a. logical form & $44.0$ & $39.7$ & $23.9$ \\
MAPO $w.$ h.a. logical form & $44.8$ & $44.3$ & $10.7$ \\
\bottomrule[1.2pt]
\end{tabular} 
}
\vspace{-0.3cm}
\caption{QA execution accuracy (\textit{EA}) on dev/test and spurious program rate of 150 samples on dev. \textit{h.a.} stands for \textit{hierarchy-aware}.}
\vspace{-0.32cm}
\label{tab:qa results}
\end{table}

\subsection{Results}
Table \ref{tab:qa results} summarizes our evaluation results.

\noindent \textbf{Weak Supervision} \indent
First, MAPO with our hierarchy-aware logical form outperforms that using its original logical form by a large margin $11.5\%$, indicating the necessity of designing a logical form leveraging hierarchies. Second, MAPO achieves the best \textit{EA} ($40.7\%$) with the lowest spurious rate ($19\%$). But $\textgreater 50\%$ questions are answered incorrectly, proving QA on HiTab is challenging. Third, though TaPas 
benefits from pretraining on tables, it
performs worse than the best logical form-based method without table pretraining.

\noindent \textbf{Partial Supervision} \indent
From Table \ref{tab:qa results}, we can conclude the effectiveness of partial supervision in two aspects. First, it improves \textit{EA}. The model learns how to deal with more cases given high-quality programs. Second, it largely lowers \textit{\%Spurious}. The model learns to generate correct programs instead of some tricks. MML, whose performance highly depends on the quality of searched programs, benefits the most ($36.7\%$ to $45.1\%$), indicating partial supervision improves the quality of consistent programs by pruning spurious ones. However, TaPas does not gain much improvements from partial supervision, which we will discuss in the next paragraph.

\noindent \textbf{Error Analysis} \indent 
For TaPas, $98.7\%$ of success cases are cell selections, which means TaPas benefits little from partial supervision. This may be caused by: (1) TaPas does not support some common operators on hierarchical table like \textit{difference}; (2) the coarse-to-fine cell selection strategy first selects columns then cells, but cells in different columns may also aggregate in hierarchical tables. 

For MAPO under partial supervision, we analyze $100$ error cases. Error cases fall into four categories: (1) entity missing ($23\%$): the header to \textit{filter} is not mentioned in question, where a common case is omitted \textit{Total}; model failure, including (2) failing to select correct regions ($38\%$) and (3) failing to generate correct operations ($20\%$); (4) out of coverage ($19\%$): question types 
unsolvable with the logical form, which is explained in Appendix~\ref{appendix:logical form}.

Spurious programs occur mostly in two patterns. In cell selection, there may exist multiple data cells with correct answers (e.g., G9,G16 in Figure~\ref{fig:realtable}), while only one is golden. In superlatives, the model can produce the target answer by operating on different regions (e.g., in both region B21:B25 and B23:B25, B23 is the largest).

\begin{figure}[t]
    \begin{center}
    \includegraphics[width=2.6in]{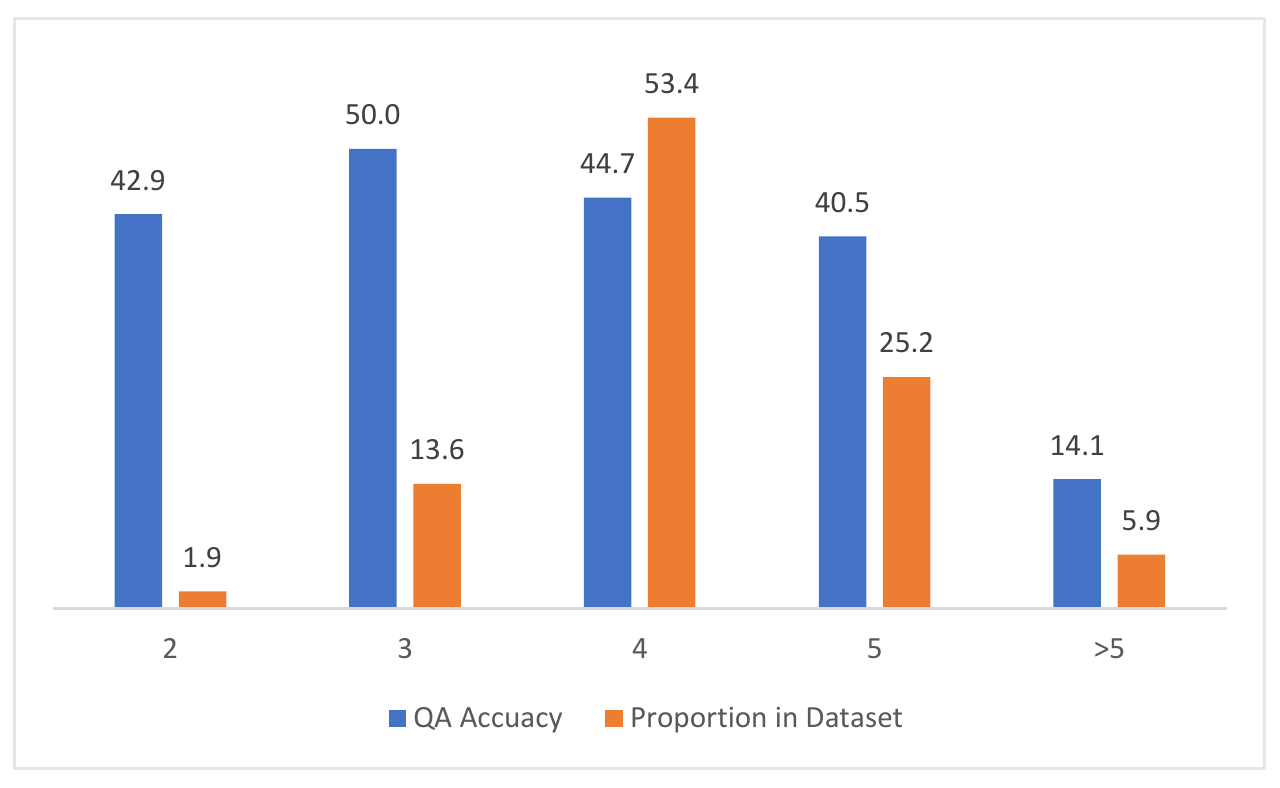}
    \end{center}
\vspace{-0.3cm}
\caption{Level-wise QA accuracy and proportion of samples with MAPO and hierarchy-aware logical form.}
\label{fig:level acc}
\vspace{-0.33cm}
\end{figure}

\noindent \textbf{Level-wise Analysis} \indent
In Figure~\ref{fig:level acc}, we present level-wise accuracy of HiTab QA with MAPO and our hierarchy-aware logical form. \textit{Level} here stands for sum of left and top header levels. As shown, the QA accuracy degrades when table level increases as table structure becomes more complex, except for level $=2$, \textit{i.e.,} tables with no hierarchies. The reason level $=2$ performs relatively worse might be that only $1.9\%$ tables without hierarchies are seen in HiTab. 
We also present an annotated table example from our dataset to illustrate in detail the challenges mentioned in Section~\ref{sec:intro} that hierarchical tables bring in Appendix~\ref{appendix:example illustrating challenges}.

% ------------------------------------------------------------- %

\section{Hierarchical Table-to-Text}\label{hitab-to-text}

\subsection{Problem Statement}\label{htt-problem-statement}

    % a general description of the data set, with clear symbols
     Some works formulate table-to-text as a summarization problem~\cite{lebret2016neural,wiseman2017challenges}. 
    However, since a full table often contains quite rich information, there lack explicit signals on what to generate, which renders the task unconstrained and the evaluation difficult.  
    On the other hand, some recent works propose \textit{controlled} generation to enable more specific and logical generation: (1) LogicNLG generates a sentence conditioned on a logical form guiding symbolic operations over given cells, but writing correct logical forms as conditions is challenging for common users who are more experienced to write natural language directly, thus restricting the application to real scenario; (2) ToTTo generates a sentence given a table 
   with a set of highlighted cells. In ToTTo's formulation, the condition of cell selection is much easier to specify than the logical form, but it neglects symbolic operations which are critical for generating some analytical sentences involving numerical reasoning in HiTab.
    
    %but specifying logical forms by users is far from a real user scenario because common people are much more experienced in writing natural language than logical forms.
    
    %is far from a real user scenario because common people are much more experienced in writing natural language than logical forms.
    
    We place HiTab as a middle-ground of ToTTo and LogicNLG to make the task more controllable than ToTTo and closer to real application than LogicNLG. In our setting, given a table, the model generates a sentence conditioned on a group of selected cells (similar to ToTTo) and operators (much easier to be specified than logical forms).
    %To assist models to produce more specific and logical descriptions, conditioned on a table, a group of cells (pointing out the target cells to describe), and operators (guiding symbolic operations on given cells and indicating analytical intents). 
    Although we use two strong conditions to guide symbolic operations over cells, there still leaves a considerable amount of content planning to be done by the model, such as retrieving contextual cells in a hierarchical table given selected cells, identifying how operators are applied on given cells, and composing sentences in a faithful and logical manner. 
    %To accurately state facts or perform operations based on user intents, extra guidance from target cells and operators can be of great help.  
    
    We now define our task as:
    given a hierarchical table $T$, highlighted cells $C$, and specified operators $O$, the goal is to generate a faithful description $S$. The dataset $H ={(T_i, S_i)}, i \in [1, N]$ is a set of $N$ table-description instances.
    Description $S_i$ is a sentence about a table $T_i$ and involves a series of operations $O_i = [O_{i1}, O_{i2}, …, O_{in}]$ on certain table cells $C_i = [c_{i1}, c_{i2}, …, c_{im}]$.
    % describe the task, how it uses each component

    %On the other hand, some data-to-text tasks simply frame data inputs in explicit meaning representations such as RDFs~\cite{nan2020dart, gardent2017webnlg}, which do not test a model’s ability to perform numerical reasoning on semi-structured context. with unknown schemas

    %We place our task at a controlled setting, where models are provided with certain guidance at generation.

    %Besides the unique hierarchical table structure and meaningful texts,
    %our task distinguishes for it owns valuable annotations of entities and quantities. They can enable more detailed and diversified attempts on table NLG.

% .2
\subsection{Controlled Generation}\label{htt-controlled-gen}

    %Full tables have sufficient yet general contents. Often by highlighting table cells~\cite{parikh2020totto} and specifying the calculation process~\cite{ibrahim2019bridging} can models produce more specific and logical generations.
    % simple explanation
    %Highlighted cells can point out the informative cells and exclude irrelevant ones.
    %Operators clarify numerical intents and reduce factual ambiguity, pushing generations beyond simple data record statements.
    % so we adopt them
    %For accurate generations towards specific user intents, we experiment with two controlled settings:
    %1) with cells of interest, and 2) further with the operators that indicate the calculation process on cells.
    
    \subsubsection{With Highlighted Cells}
        An entity or quantity in text can be supported by table cells if it is directly stated in cell contents, or can be logically inferred by them. 
        %Motivated by ~\cite{parikh2020totto}, cell highlights help models to produce more specific generations.  
        Different from only taking data cells as highlighted cells~\cite{parikh2020totto}, we also take header cells as highlighted cells, and it is usually the case for superlative ARG-type operations on a specific header level in hierarchical tables, e.g., ``Teaching assistantships'' is retrieved by ARGMAX in Figure~\ref{fig:realtable}. 
        In our dataset, highlighted cells are extracted from annotations of the entity and quantity alignment. %, while in practice, we hope that highlighted cells can be flexibly selected based on user interests.

    \subsubsection{With Operators}
        Highlighted cells can tell the target for text generation, but is not sufficient, especially for analytical descriptions involving cell operations in HiTab. 
        %Some works use logical forms~\cite{chen2020logic2text} or mathematical expressions~\cite{ibrahim2019bridging} to ground quantities with their calculation process. It motivates us to use formulas as additional controls for text generation. 
        %Different from logical forms that are hard to write by common users~\cite{chen2020logical}, we propose to use operators as conditions that are very easy to specify by users.
        So we propose to use operators as extra control. It contributes to text clarity and meaningfulness in two ways.
        % first contrib
        (1) It clarifies the numerical reasoning intent on cells. 
        For example, given the same set of data cells, applying SUM, AVERAGE, or COUNT conveys different meanings thus should yield different texts.
        % second contrib
        (2) Operation results on highlighted cells can be used as additional input sources. Existing seq2seq models are not powerful enough to do arithmetic operations~\cite{thawani2021representing}, e.g., adding up a group of numbers, and it greatly limits their ability to generate correct numbers in sentences. Explicitly pre-computing the calculation results is a promising alternative way to mitigate this gap in seq2seq models. Operators are extracted from annotations of formulas shown in Table~\ref{tab:formulas}.
        
        %Even with these controls, text generation on hierarchical tables is still a challenge due to the complex hierarchical indexing and implicit semantic relationships among cells.
        
    \subsubsection{Sub Table Selection and Serialization} 
    \indent 
    
    \textbf{Sub Table Selection} \indent 
    Under controls of selected cells and operators, we devise a heuristic to retrieve all contextual cells as a sub table. 
    (1) We start with highlighted cells extracted from our entity and quantity alignment, then use the extracted table hierarchy to group the selected cells into the top header, the left header, and the data region.
    (2) Based on the extracted table hierarchy, we use the source set of top and left header cells to include their indexed data cells, and we also use the source set of data cells to include corresponding header cells. 
    (3) We also include their parent header cells in table hierarchy to construct a full set of headers. 
    In the end, we take the union of of them as the result of sub table selection.

    \textbf{Serialization} \indent 
    On each sub table, we do a row-turn traversal on linked cells and concatenate their cell strings using [SEP] tokens. Operator tokens and calculation results are also concatenated with the input sequence. 
    We also experimented with other serialization methods, such as header-data pairing or template-based method, yet none reported superiority over the simple concatenation. Appendix~\ref{appendix:nlg case} gives an illustration.

\subsection{Experiments}\label{htt-experiments}

%\subsubsection{Baseline} %\label{htt-exp-baseline}
    We conduct experiments by fine-tuning four state-of-the-art text generation methods on HiTab.
    % including Pointer Generator~\cite{see2017get}, BERT-to-BERT~\cite{rothe2020leveraging}, BART~\cite{lewis2019bart}, and T5~\cite{raffel2019exploring}. For space limit, we leave the details of baselines and our implementations in Appendix~\ref{appendix:nlg baseline} for interested readers.
    
    \noindent \textbf{Pointer Generator}~\cite{see2017get} \indent
    A LSTM-based seq2seq model with copy mechanism. While originally designed for text summarization, it is also used in data-to-text~\cite{gehrmann2018end}.
    %The model uses two-layer bi-directional LSTMs for the encoder with $300$-dim word embeddings and $300$ hidden units. We perform fine-tuning using batch size $2$, learning rate $0.05$, and beam size $5$.
    
    \noindent \textbf{BERT-to-BERT}~\cite{rothe2020leveraging} \indent
    A transformer encoder-decoder model~\cite{vaswani2017attention} initialized with BERT~\cite{devlin2018bert}.
    %by loading the checkpoint named `bert-base-uncased' provided by the huggingface/transformers repository. We perform fine-tuning using batch-size $2$ and learning rate $3e^{-5}$.
    
    \noindent \textbf{BART}~\cite{lewis2019bart} \indent
    A pre-trained denoising autoencoder with standard Transformer-based architecture and shows effectiveness in NLG. %We align model configuration with the BASE version of BART, and use the model `facebook/bart-base' in huggingface/transformers. During fine-tuning, we use a batch size of $8$ and a learning rate of $2e^{-4}$.
    
    \noindent \textbf{T5}~\cite{raffel2019exploring} \indent
    A transformer-based pre-trained model. It converts all textual language problems into text-to-text and proves to be effective. 
    %We use the pre-trained model `t5-base' in huggingface/transformers. For fine-tuning, we set batch size to $8$ and learning rate to $2e^{-4}$.

\subsubsection{Evaluation Metrics}

    We use two automatic metrics, BLEU and PARENT.~BLEU~\cite{papineni2002bleu} is broadly used to evaluate text generation. PARENT~\cite{dhingra2019handling} is proposed specifically for data-to-text evaluation that additionally aligns n-grams from the reference and generated texts to the source table. 

\subsubsection{Experiment Setup}
    % dataset split
    Samples are split into train ($70\%$), dev ($15\%$), and test ($15\%$) sets just the same as the QA task.
    %, \textit{i.e.}, samples of a table always appear in the same split.
    The maximum length of input/output sequence is set to $512$/$64$. Implementation details of all baselines are given in Appendix~\ref{appendix:nlg baseline}.

\subsubsection{Experiment Result and Analysis}

    \begin{table}
        \resizebox{7.68cm}{!}{
        \begin{tabular}{l c p{2cm}<{\centering}  c c}
        \toprule[1.2pt]
            \vspace{0.05cm}
            % \multirow{3}{*}{\textbf{Model}} & \multicolumn{4}{c}{\textit{Controlled Settings}} \\
            \vspace{0.1cm}
            % \cline{2-5}
            \multirow{2}{*}{\textbf{Model}} & \multicolumn{2}{c}{\textbf{Cell Highlight}}  & \multicolumn{2}{c}{\textbf{Cell \& Calculation}}\\
                {} & {BLEU-4} & {PARENT} & {BLEU-4} & {PARENT} \\
            \hline
            {Pointer-Generator} & ${5.8}$ & ${8.8}$ & ${9.0}$ & ${10.8}$ \\
            {BERT-to-BERT} & ${11.4}$ & ${16.7}$ & ${11.7}$ & ${15.4}$ \\
            {BART} & ${17.9}$ & ${28.0}$ & ${23.8}$ & ${31.4}$ \\
            {T5} & ${\textbf{19.5}}$ & ${\textbf{35.7}}$ & ${\textbf{26.6}}$ & ${\textbf{36.9}}$ \\
            \bottomrule[1.2pt]
        \end{tabular}
        }
        \vspace{-0.3cm}
        \caption{Results of hierarchical table-to-text.}
        \label{tab:htt-res-link}
        \vspace{-0.32cm}
    \end{table}

    % overall difficult
   As shown in Table~\ref{tab:htt-res-link}, \textbf{first}, from an overall point of view, both metrics are not scored high. This well proves the difficulty of HiTab. It could be caused by the hierarchical structure, as well as statements with logical and numerical complexity.
    % between model baselines
    \textbf{Second}, by comparing two controlled scenarios (cell highlights \& both cell highlights and operators), we see that adding operators to conditions greatly help models to generate descriptions with higher scores, showing the effectiveness of our augmented conditional generation setting.
    %So, to produce texts in specific intents, the more controlled input is, the more meaningful a generated sentence can be.
    \textbf{Third}, results on two controlled scenarios across baselines are quite consistent. 
    Replacing the traditional LSTM with transformers shows large increasing.
    Leveraging seq2seq-like pretraining yields a rise of $+6.5$ BLEU and $+11.3$ PARENT.
    Lastly, between pretrained transformers, T5 reports higher scores over BART, probably for T5 is more extensively tuned during pre-training.
    % better using opertors

    % by hierarchy depths
    Further, to study the generation difficulty concerning \textbf{table hierarchy}, we respectively evaluate samples at different hierarchical depths, \textit{i.e.}, table's maximum depths in top and left header trees. 
    In groups of 2, 3, 4+ depth, BLEU scores $31.7$, $26.5$, and $21.3$; PARENT scores $40.9$, $36.5$, and $31.6$.
    The reason could be that, as the table header hierarchy grows deeper, the data indexing becomes increasingly compositional, rendering it harder to baseline models to configure entity relationships and compose logical sentences.

% ------------------------------------------------------------- %

\section{Related Work}

\textbf{Table-to-Text} Existing datasets are restricted in flat tables or specific subjects~\cite{liang2009learning,chen2008learning,wiseman2017challenges,novikova2016crowd,banik2013kbgen,lebret2016neural,moosavi2021learning}.
The most related table-to-text dataset to HiTab is ToTTo~\cite{parikh2020totto}, in which complex tables are also included.
There are two main differences between HiTab and ToTTo: (1) in ToTTo, hierarchical tables only account for a small proportion ($5\%$), and there are no indication and usage of table hierarchies. (2) in addition to cell highlights, Hitab conditions on operators that reflect symbolic operations on cells.
%. In contrast, hierarchies are explicitly extracted and studied in HiTab for public usage. %Another highly

\textbf{Table QA} mainly focuses on DB tables~\cite{wang2015building,yu2018spider,zhong2017seq2sql} and flat web tables~\cite{pasupat2015compositional,sun2016table}. Recently, there are some datasets on domain-specific table QA~\cite{chen2021finqa,zhu2021tat} and jointly QA over tables and texts~\cite{chen2020hybridqa,zhu2021tat}, but hierarchical tables still have not been studied in depth. CFGNN~\cite{zhang2020cfgnn} and GraSSLM~\cite{zhang2020graph} uses gragh neural networks to encode tables for QA, but all tables are database tables and relational web tables without hierarchies, respectively. ~\citet{wang2021retrieving} include some hierarchical tables but only focuses on table search.

%There exist two popular methodologies, logical form-based semantic parsing\cite{mapo,liang2017neural,yin2020tabert} and end-to-end parsing without logical form~\cite{herzig2020tapas}.

%\textbf{Table structure understanding}
%involves a series of tasks: table detection~\cite{dong2019tablesense}, table recognition~\cite{nishida2017understanding,ghasemi2018tabvec}, hierarchy extraction~\cite{chen2014integrating,wang2020structure}, cell classification~\cite{gol2019tabular,pujara2021hybrid,dong2019semantic}, etc. By stringing them together, ~\cite{chen2013automatic,koci2019xlindy,kardas2020axcell} explored extracting relational data from semi-structured tables, but need human interactions to get precise results.

\begin{figure}[t]
    \begin{center}
    \includegraphics[width=2.9in]{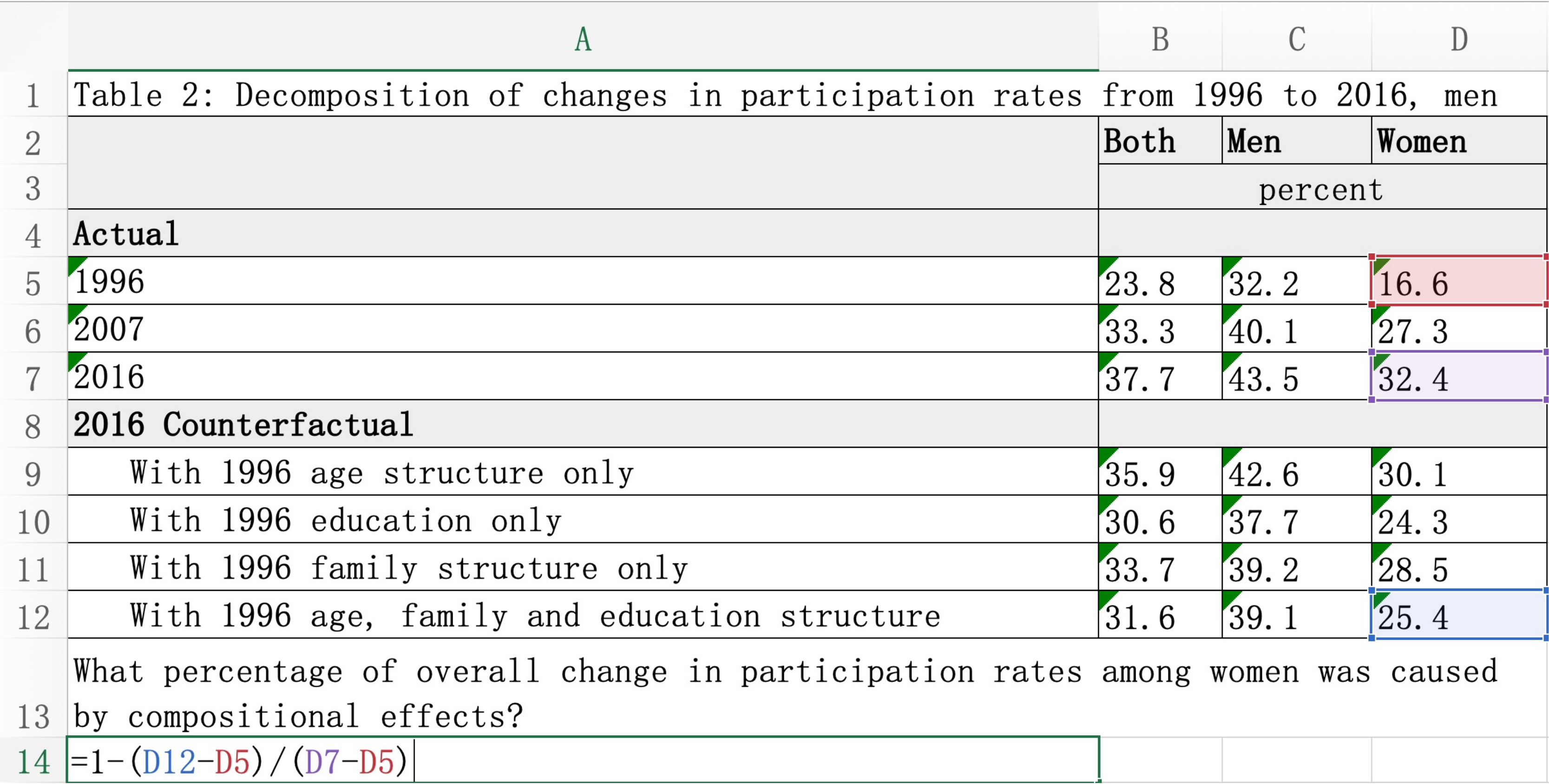}
    \end{center}
\caption{A meaningful but challenging case in HiTab. }
\label{fig:badcase}
\end{figure}

\section{Discussion}
HiTab also presents cross-domain and complicated-calculation challenges.
(1) To explore cross-domain generalizability, we randomly split train/dev/test by domains for three times and present the average results of our best methods in Table \ref{tab:domain results}. We found decreases in all metrics in QA and NLG.
(2) Figure~\ref{fig:badcase} shows a case that challenges existing methods: performing complicated calculations requires to jointly consider quantity relationships, header semantics, and hierarchies. 

\begin{table}[t]
%\vspace{-0.3cm}
\centering
\small
\begin{tabular}{l c c c}
\toprule[1.2pt]
\vspace{0.05cm}
\textbf{Method}  & \multicolumn{2}{c}{\textbf{Test Accuracy}} \\
% \cdashline{2-3}[2pt/1pt]
\vspace{0.05cm}
MAPO $w.$ partial supervision & \multicolumn{2}{c}{$32.6$}\\
\hline
\vspace{0.05cm}
~                & \textbf{BLEU} & \textbf{PARENT} \\
% \cdashline{2-3}[2pt/1pt]
\;T5 $w.$ cell \& calculation & ${16.9}$ & ${28.8}$ \\
\bottomrule[1.2pt]
\end{tabular}
\caption{Results of cross-domain evaluation.}

\label{tab:domain results}
\end{table}

\section{Conclusion}
We present a new dataset, HiTab, that simultaneously supports QA and NLG on hierarchical tables, where tables are collected from statistical reports and Wikipedia in various domains. Importantly, we provide fine-grained annotations on entity and quantity alignment. In experiments, we introduce strong baselines and conduct detailed analysis on QA and NLG tasks on HiTab. Results suggest that HiTab can serve as a challenging and valuable benchmark for future research on complex tables.

% ------------------------------------

\section{Ethical Considerations} \label{appendix:ethics}
This work presents HiTab, a free and open English dataset for the research community to study table question-answering and table-to-text over hierarchical tables. Our dataset contains well-processed tables, annotations (QA pairs, target text, and bidirectionally mappings between entities and quantities in text and the corresponding cells in table), recognized table hierarchies, and source code. Data in HiTab are collected from two public organizations, StatCan and NSF. Both of them allow sharing and redistribution of their public reports, so there is no privacy issue. We collect tables and accompanied descriptive sentences from StatCan and NSF. We also include hierarchical tables in Wikipedia from ToTTo, which is a public dataset under MIT license, so there is no risk to use it. And in the labeling process, annotators need to check if there exist any names or uniquely identifies individual people or offensive content. They did not find any such sensitive information in our dataset. We recruit $18$ students or graduates in computer science, finance, and English majors from top universities($13$ females and $5$ males). Each student is paid $\$7.8$ per hour (above the average local payment of similar jobs), totally spending $2,400$ hours. We finally get $3,597$ tables and $10,672$ well-annotated sentences. And the data got approval from an ethics review board by an anonymous IT company. The details for our data collection and characteristics are introduced in Section ~\ref{sec:dataset}.

\bibliographystyle{acl_natbib}
\bibliography{acl_natbib}
        
\clearpage
\appendix
% \onecolumn

% -------------------------------------------------------------------------------------
\section{More Details on Dataset}
\subsection{Dataset Preprocessing} \label{appendix:dataprepare}
We filter tables using these constraints:  (1) number of rows and columns are more than $2$ and less than $64$; (2) cell strings have no more than one non-ASCII character and $20$ tokens; (3) hierarchies are successfully parsed via the method in \ref{subsec:hierarchy extraction}. (4) hierarchies have no more than four levels on one side. Finally, $85\%$ tables meet all constraints.

% ------------------------------------
\subsection{Annotation Artifacts} \label{appendix:annotation artifacts}
Annotation artifacts are common in large scale NLP datasets, which may raise unwanted statistical correlations making the task easier~\cite{gururangan2018annotation}. In HiTab, the annotation artifacts may come from homogeneous patterns of questions. To address this issue, we ask annotators to revise questions from the high-quality descriptions from statistical reports from $28$ domains to guarantee the diversity and naturalness, and encourage them to choose the best way to raise question reflecting the overall meaning of the description. 
To further check whether and where artifacts may exist in our dataset, we conduct two experiments on QA and count the ratio of answer occurring in the question:
\begin{itemize}
    \item Use table as only input without question, to see if there is a potential pattern between table and answer. We train BERT+MAPO for $10,000$ steps and TaPas for $10$ epochs. Both methods can't converge under this setting, with $4.0\%$ and $2.6\%$ accuracy on the test set. The poor performance indicates model can't learn the answers by exploring and leveraging artifacts between the table and answer, and thus should learn to jointly inference the question and table.
    \item Shuffle the rows and columns of table randomly. Experiments show similar performance~($\pm 1\%$) between our original tables and shuffled tables. The result shows that the correlation between answer and table cell position is very little, thus model can't choose some specific positions, e.g., cell at the first row and first column, as a shortcut prediction.
    \item The ratio that answer occurs in the question is only $5.3\%$. Model that only learns to retrieve the question can't achieve high performance.
\end{itemize}

% ------------------------------------
\subsection{Domain Distribution} \label{appendix:domain distribution}
\begin{figure}[htp]
    \begin{center}
    \includegraphics[width=3.15in]{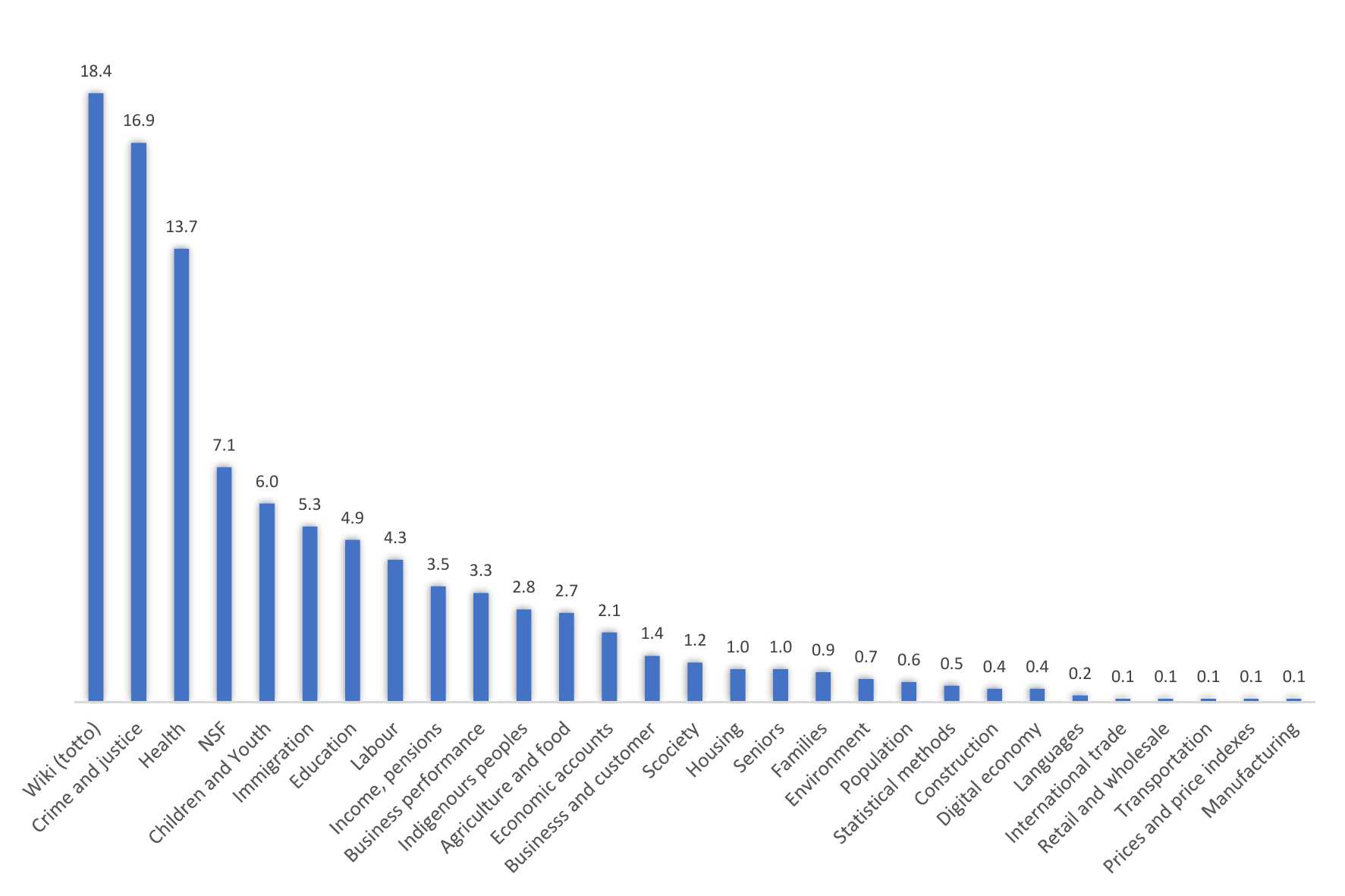}
    \end{center}
\caption{Proportion of samples in different $29$ domains.}
\label{fig:domain distribution}
\end{figure}

The full $29$ domains of sample distribution in HiTab are shown in Figure~\ref{fig:domain distribution}.

% ------------------------------------
\subsection{Annotation Interface}
\begin{figure*}[t]
\begin{center}
\includegraphics[width=6in]{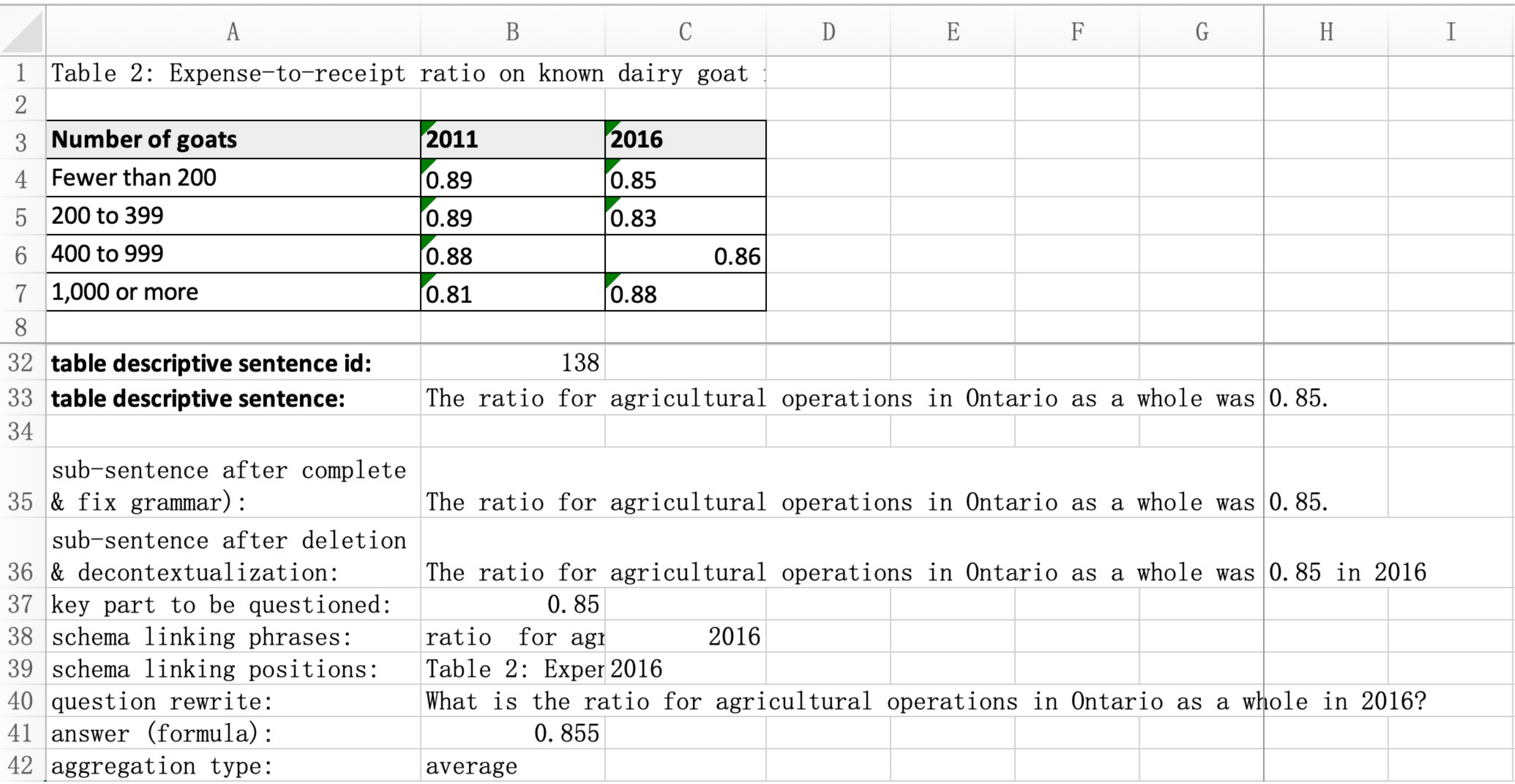}
\end{center}
\caption{Annotation interface in Excel.}
\label{fig:annotation interface}
\end{figure*}

The annotation interface looks like Figure~\ref{fig:annotation interface}. Since spreadsheet formula is easy to write, execute, and check, the spreadsheet is naturally a great annotation tool. Annotators can use the Excel formula conveniently for cell linking and calculation in entity alignment and answering questions.

% -------------------------------------------------------------------------------
\section{Hierarchical Table-to-Text}

\subsection{Illustration on Controlled Generation in Hierarchical Table-to-Text.} \label{appendix:nlg case}
\begin{figure}[p]
    \begin{center}
    \includegraphics[width=3.2in]{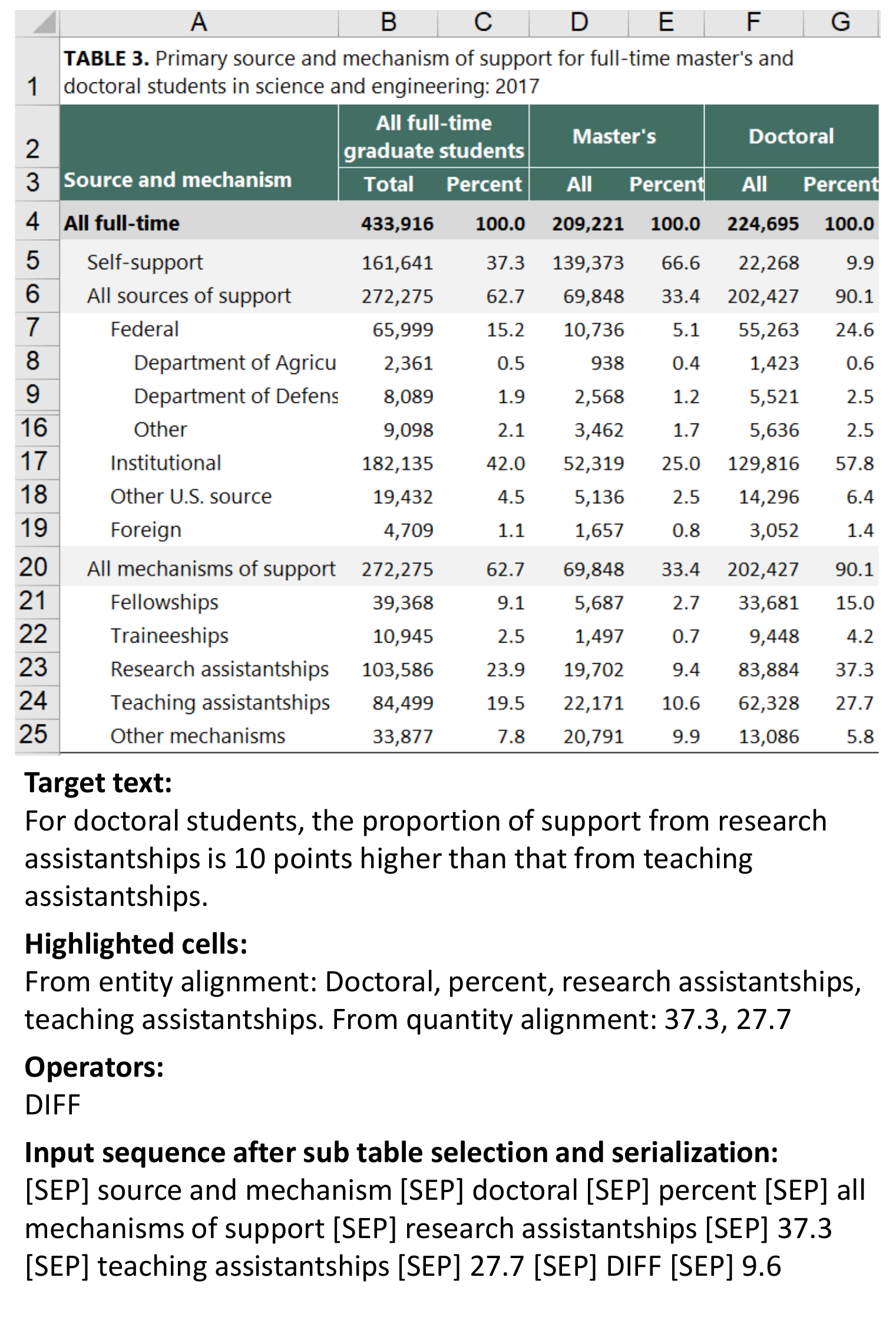}
    \end{center}
    \vspace{-0.3cm}
\caption{An illustration on controlled generation. }
\label{fig:nlg case}
\vspace{-0.32cm}
\end{figure}

Please find the illustration shown in Figure~\ref{fig:nlg case}.

\subsection{Baseline Implementation Details} \label{appendix:nlg baseline}

We perform optimized tuning for baselines using the following settings.

\noindent \textbf{Pointer Generator}~\cite{see2017get} \indent
    A LSTM-based seq2seq model with copy mechanism. 
    The model uses two-layer bi-directional LSTMs for the encoder with $300$-dim word embeddings and $300$ hidden units. We perform fine-tuning using batch size $2$, learning rate $0.05$, and beam size $5$.
    
    \noindent \textbf{BERT-to-BERT}~\cite{rothe2020leveraging} \indent
    A transformer encoder-decoder model~\cite{vaswani2017attention} where the encoder and decoder are both initialized with BERT~\cite{devlin2018bert} by loading the checkpoint named `bert-base-uncased' provided by the huggingface/transformers repository. We perform fine-tuning using batch-size $2$ and learning rate $3e^{-5}$.
    
    \noindent \textbf{BART}~\cite{lewis2019bart} \indent
    BART is a pre-trained denoising autoencoder for seq2seq language modeling. It uses standard Transformer-based architecture and shows effectiveness in NLG. We align model configuration with the BASE version of BART, and use the model `facebook/bart-base' in huggingface/transformers. During fine-tuning, we use a batch size of $8$ and a learning rate of $2e^{-4}$.
    
    \noindent \textbf{T5}~\cite{raffel2019exploring} \indent
    T5 is also a transformer-based pre-training LM. It trains extensively on text-to-text tasks and scores high on generation tasks. We use the pre-trained model `t5-base' in huggingface/transformers. For fine-tuning, we set batch size to $8$ and learning rate to $2e^{-4}$.

We use a beam size of $5$ to search decoded outputs (sequence lengths range from $8$ to $60$ tokens)

\begin{table*}[htp]

\begin{center}
\scalebox{0.92}{
\begin{tabular}{l l l l}
\toprule[1.5pt]
\textbf{Function}  & \textbf{Arguments} & \textbf{Returns}  & \textbf{Description} \\
\hline
(\textbf{filter\_tree} h)   &  \textbf{h}: a header  & a region   & Select a region indexed by sub-tree of \\
~                             &  ~  & ~          & the given header in the given region. \\
\hline
(\textbf{filter\_level} l)  & \textbf{l}: a level   & a region   & Select a region indexed by headers on \\
~                             &  ~   & ~          & the given level in the given region. \\
\hline
(\textbf{argmax} k)         & \textbf{k}: a number   & a list of headers  & Find the header(s) with k-th largest/  \\
(\textbf{argmin} k)         & ~   & ~                  & smallest value in the region. [Input region  \\
~                             & ~                      & ~                  &  should have one row or one column of data]\\
\hline
(\textbf{max} l)            & \textbf{l}: a level   & a region           & Maximum/minimum/sum/average of the given \\
(\textbf{min} l)            & ~    & ~                       & region, grouping by headers of the given level,  \\
(\textbf{sum} l)            & ~                      & ~                       & \textit{i.e.}, data values aggregate according to their \\
(\textbf{average} l)        & ~                      & ~                       & header strings on the given level. \\
\hline
(\textbf{count} l)           & \textbf{l}: a level    & a number                      &  Count the number of headers on the given\\
~                             &  ~   & ~                       &  level of given region.\\
\hline
(\textbf{difference})       & ~   & a number  & Absolute difference, proportion and     \\
(\textbf{proportion})       & ~   & ~  & difference rate of given two elements \\ 
(\textbf{proportion\_rev})  & ~   & ~  &  $a$ and $b$ in region. $rev$ means changing  \\
(\textbf{difference\_rate})       & ~ & ~  & order of operands. e.g., $proportion$ applies \\
(\textbf{difference\_rate\_rev})       & ~ & ~  & $b/a$ and $proportion\_rev$ applies $a/b$. \\
~      & ~ & ~  & [Input region should have two data elements] \\
\hline
(\textbf{greater\_than} n)  & \textbf{n}: a number & a list of headers &  Find the header(s) with data value(s) that have\\
(\textbf{greater\_eq\_than} n)  & ~ & ~ & certain order relation with the given number.\\
(\textbf{less\_than} n)  & ~ & ~ & [Input region should have one row or one \\
(\textbf{less\_eq\_than} n)  & ~ & ~ & column of data]\\
(\textbf{eq} n)  & ~ & ~ & \\
(\textbf{not\_eq} n)  & ~ & ~ & \\
\hline
(\textbf{opposite}) & ~ &  a number & Take opposite value of data in a given region. \\ 
~ & ~  &  ~ & [Input region should have one data element] \\ 
\bottomrule[1.5pt]
\end{tabular}
}
\end{center}   
\caption{Function list of hierarchy-aware logical form }
\label{tab:logical form}
\end{table*}

% -------------------------------------------------------------------------------------
\section{Hierarchical Table QA}
\subsection{Logical Form Function List} \label{appendix:logical form}
We list our logical form functions in Table~\ref{tab:logical form}.

Union selection is required for comparative and arithmetic operations. It is achieved by allowing variable number of headers in $filter\_tree$, where ``variable'' is one or two in practice.

In our implementation, a function by default takes the selected region of last function as input region to prune search space. We use grammars to filter left headers before top headers, and a  $(filter\_level)$ is necessary after filtering one direction of tree even when only the leaf level is available. And we deactivate order relation functions (e.g., \textit{eq} function) and the order argument \textit{k} in \textit{argmax/argmin} because there are few questions in these types and activating them will largely increase number of spurious programs when searching. 

The logical form coverage after deactivation is $78.3\%$ in $300$ iterations of random exploration. Some typical question types that can not be covered are: (1) scale conversion, e.g., $0.984$ to $98.4\%$, (2) operating data indexed by different levels of headers, e.g., proportion of total, (3) complex composite operations, e.g., Figure~\ref{fig:badcase}.

\begin{table}[t]
\begin{center}
\scalebox{0.85}{
\begin{tabular}{l l}
\toprule[1.5pt]
\textbf{Question}  & \textbf{Logical Forms}  \\
\hline
\textbf{Cell Selection}                   & (filter\_tree \;2012)  \\
Q: What is the GDP                        & (filter\_tree  \;china) \\
  \;\;\;\;\;of China in 2012?                      & (filter\_level \;LEFT\_2) \\
~                                         & (filter\_tree \;gdp) \\
~                                         & (filter\_level \;TOP\_1) \\
\hline
\textbf{Superlative}                           & (filter\_tree \;2012) \\
Q: Which country has                           & (filter\_level \;LEFT\_2) \\
  \;\;\;\;\;the highest GDP in 2012?                    & (filter\_tree \;gdp) \\
~                                              & (filter\_level \;TOP\_1) \\
~                                              & (argmax \;1) \\
\hline
Q: How much more is    & (filter\_tree \;u.s. \;china) \\
\;\;\;\;\;U.S. GDP higher than                                                        & (filter\_level \;LEFT\_2) \\
\;\;\;\;\;China in 2013?                                                       & (filter\_tree \;gdp) \\
~                                                        & (filter\_level \;TOP\_1) \\
~                                                        & (difference) \\
\bottomrule[1.5pt]
\end{tabular}
}
\end{center}   
\caption{Examples of our logical form. The table to be questioned is in Fig.~\ref{fig:linearization}.  \textit{LEFT\_1} is a symbol for the first level on the left. }
\label{tab:demo logical forms}
\end{table}

% ------------------------------------
\subsection{Examples of Logical Form Execution} \label{appendix:qa examples}
Take the table in Figure~\ref{fig:linearization} as input table, we demonstrate three types of questions with complete logical forms in Table~\ref{tab:demo logical forms}.

% -------------------------------------
\subsection{Pruning Rules in Searching} \label{appendix:trigger words}
We use trigger words and POS tags for some functions in random exploration, which is inspired by \cite{zhang2017macro, mapo}. Functions are allowed to be selected only when triggers appear in the question. Triggers are listed in Table~\ref{tab:trigger words}.

\begin{table}[t]
\begin{center}
\begin{tabular}{l l}
\toprule[1.5pt]
\textbf{Function}  & \textbf{Trigger Words}  \\
\hline
argmax  & JJR, JJS, RBR, RBS, top,   \\
argmin & first, bottom, and last. \\
\hline
max & JJS, RBS \\
min & ~ \\
\hline
average & average, mean \\
\hline
sum & all, combine, total, sum \\
\hline
count & how, many, total, number \\
\hline
difference & difference, more, than, \\
difference\_rate &  change,compare, JJR \\
difference\_rate\_rev & RBR.\\
\hline
proportion & times, percent, \\
proportion\_rev & percentage, fraction  \\
\bottomrule[1.5pt]
\end{tabular}
\end{center}   
\caption{Trigger Words for Functions}
\label{tab:trigger words}
\end{table}

% -------------------------------------
\subsection{Table Linearization} \label{appendix:linearization}
\begin{figure*}[p]
    \begin{center}
    \includegraphics[width=6.4in]{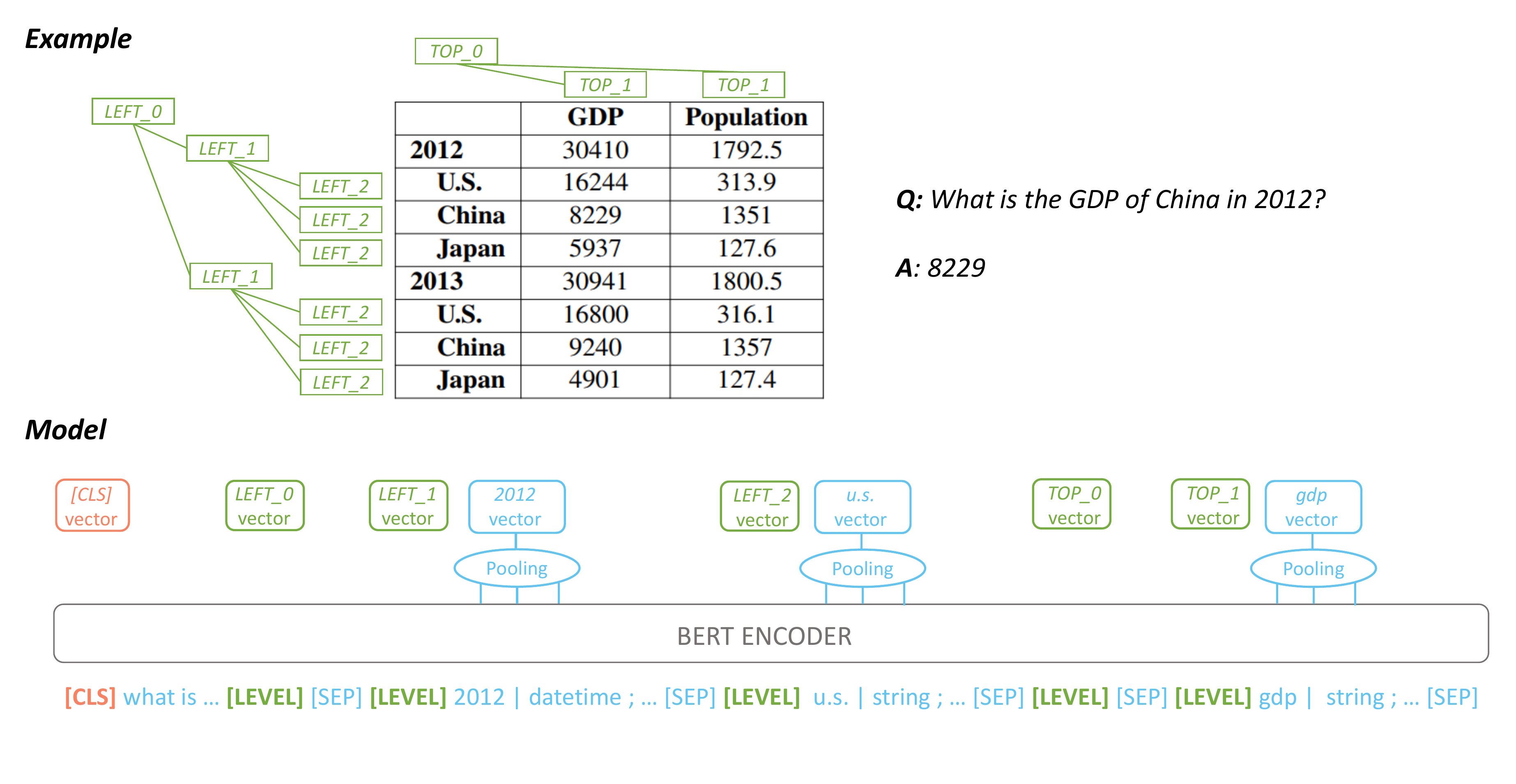}
    \end{center}
\caption{An QA example table with hierarchy and its linearized input to the encoder. Each level in the hierarchical header starts with a \textit{LEVEL} token to learn a level representation. \textit{LEFT\_k} means the \textit{k}th level in the left tree. Each header cell has a unique header cell representation. }
\label{fig:linearization}
\end{figure*}

We linearize the question and table according to Figure~\ref{fig:linearization}.

The input is concatenation of question and table. Table is linearized by putting headers in level order. Each level is led by a \textit{[LEVEL]} token to gather current level embedding. The first \textit{[LEVEL]} token stands for level zero of left. Each header is linearized as \textit{name} $|$ \textit{type}. \textit{name} is the tokenized header string. \textit{type} is the entity type parsed by Stanford CoreNLP, which includes ``string'', ``number'', ``datetime'' in our case. Headers with the same \textit{name} will gather token embeddings by mean pooling.

% -------------------------------------
% \subsection{More Experiment Results} \label{appendix:more qa results}
% Here we present more detailed error analysis on TaPas and our method.

% For TaPas, $98.7\%$ of success cases are cell selections, which means TaPas benefits little from partial supervision. This may be caused by: (1) TaPas does not support some common operators on hierarchical table like \textit{difference}; (2) the coarse-to-fine cell selection strategy first selects columns then cells, but cells in different columns may also aggregate in hierarchical tables. 

% For MAPO under partial supervision, we analyze $100$ error cases. Error cases fall into four categories: (1) entity missing ($23\%$): the header to \textit{filter} is not mentioned in question, where a common case is omitted \textit{Total}; model failure, including (2) failing to select correct regions ($38\%$) and (3) failing to generate correct operations ($20\%$); (4) out of coverage ($19\%$): question types 
% unsolvable with the logical form, which is explained in Appendix~\ref{appendix:logical form}.

% Spurious programs occur mostly in two patterns. In cell selection, there may exist multiple data cells with correct answers (e.g., G9,G16 in Figure~\ref{fig:realtable}), while only one is golden. In superlatives, the model can produce the target answer by operating on different regions (e.g., in both region B21:B25 and B23:B25, B23 is the largest).

% ------------------------------------
\subsection{Illustration on Challenges in Hierarchical Table} \label{appendix:example illustrating challenges}
\begin{figure*}[p]
    \begin{center}
    \includegraphics[width=6.4in]{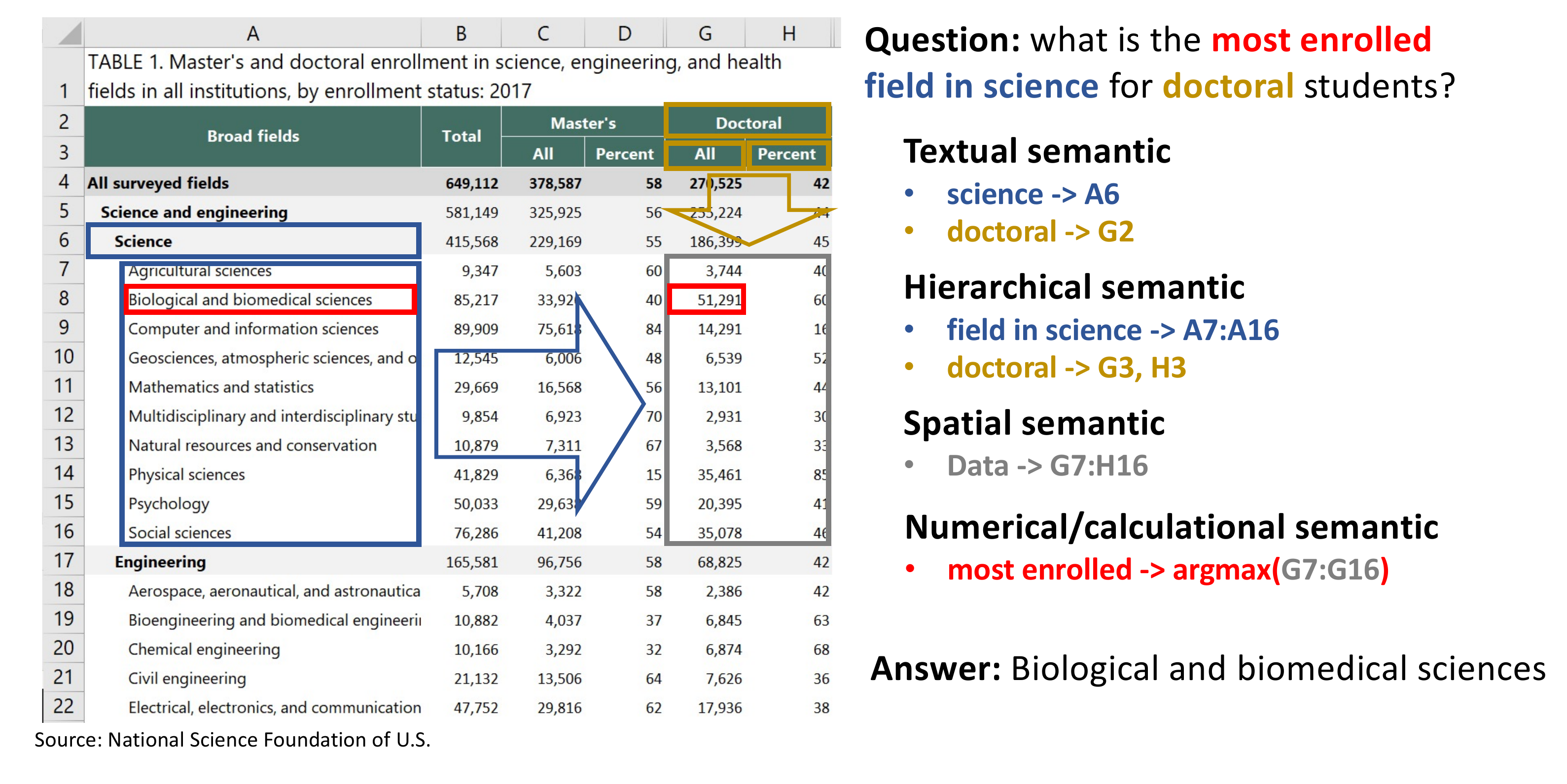}
    \end{center}
\caption{A detailed annotated example to illustrate challenges in hierarchical table.}
\label{fig:example illustrating challenges}
\end{figure*}

We present an annotated example in Figure~\ref{fig:example illustrating challenges} to show the challenges of hierarchical table introduced in Section~\ref{sec:intro}. 

To precisely answer the question in the figure, the model/method first needs to hierarchically index the grey region with ``field in science" and ``doctoral", which requires understanding of textual and spatial semantics of the hierarchical table since the textual headers are spatially~(seen as a tree) related with the region. Second, from the phrase ``most enrolled", it should further indexes ``All"~(column G) rather than ``Percent"~(column H) and infers \textit{argmax} operation, , which calls for the ability to distinguish between different calculation relationships.

% ~\footnotetext{https://www.nsf.gov/statistics/2019/nsf19319/}

\end{document}